\documentclass[letterpaper,twocolumn,10pt,english]{article}
\usepackage{usenix2019_v3}

\usepackage{tikz}
\usepackage{amsmath}
\usepackage{graphicx}
\usepackage{epstopdf}
\usepackage{filecontents}
\usepackage{caption}
\usepackage{multirow}
\usepackage[T1]{fontenc}
\usepackage{array}
\makeatletter

\providecommand{\tabularnewline}{\\}

\makeatother
\usepackage{babel}
\usepackage{color}

\begin{document}

\date{}

\title{\Large \bf AccUDNN: A GPU Memory Efficient Accelerator for Training \\
    Ultra-deep Neural Networks}

\author{
{\rm Jinrong Guo$^{1,2}$, Wantao Liu$^{1}$*, Wang Wang$^{1,2}$, Qu Lu$^{1,2}$,
\rm Songlin Hu$^{1}$, Jizhong Han$^{1}$, Ruixuan Li$^{1}$} \\
\rm Institute of Information Engineering, Chinese Academy of Sciences$^{1}$, \\
\rm School of Cyber Security, University of Chinese Academy of Sciences$^{2}$ \\
{\rm \{\emph{guojinrong, liuwantao, wangwang, luqu, husonglin, hanjizhong, liruixuan}\}@iie.ac.cn}
}


\maketitle

\begin{abstract}
Typically, Ultra-deep neural network(UDNN) tends to yield high-quality model, but its training process is usually resource intensive and time-consuming. Modern GPU's scarce DRAM capacity is the primary bottleneck that hinders the trainability and the training efficiency of UDNN. In this paper, we present "AccUDNN", an accelerator that aims to make the utmost use of finite GPU memory resources to speed up the training process of UDNN. AccUDNN mainly includes two modules: memory optimizer and hyperparameter tuner. Memory optimizer develops a performance-model guided dynamic swap out/in strategy, by offloading appropriate data to host memory, GPU memory footprint can be significantly slashed to overcome the restriction of trainability of UDNN. After applying the memory optimization strategy, hyperparameter tuner is designed to explore the efficiency-optimal minibatch size and the matched learning rate. Evaluations demonstrate that AccUDNN cuts down the GPU memory requirement of ResNet-152 from more than 24GB to 8GB. In turn, given 12GB GPU memory budget, the efficiency-optimal minibatch size can reach 4.2x larger than original Caffe. Benefiting from better utilization of single GPU's computing resources and fewer parameter synchronization of large minibatch size, 7.7x speed-up is achieved by 8 GPUs' cluster without any communication optimization and no accuracy losses.
\end{abstract}

\section{Introduction}
\newcommand\blfootnote[1]{%
\begingroup
\renewcommand\thefootnote{}\footnote{#1}%
\addtocounter{footnote}{-1}%
\endgroup
}
\blfootnote{* Corresponding author}
Scale of data volume and computation infrastructure together make current deep learning flourish, especially in computer vision field\cite{he2016deep,krizhevsky2012imagenet,TP-toolbox-web,simonyan2014very,smith2016gradual}. From 8-layer AlexNet\cite{krizhevsky2012imagenet} to 152-layer ResNet\cite{he2016deep}, the neural network architecture is getting deeper and the corresponding model quality is getting better. However, training a hundreds-of-layers UDNN, as ResNet, is a resource intensive task, not only involving a significant amount of computing resources but also memory space. Nowadays, GPU's multi-core parallel architecture along with high memory bandwidth makes it a popular choice to train deep neural network model, but regrettably, the limited size of GPU DRAM capacity, generally at most 24GB, is far from sufficient to accommodate UDNN.
\begin{figure}[ht]
\centering
\includegraphics[scale=0.2]{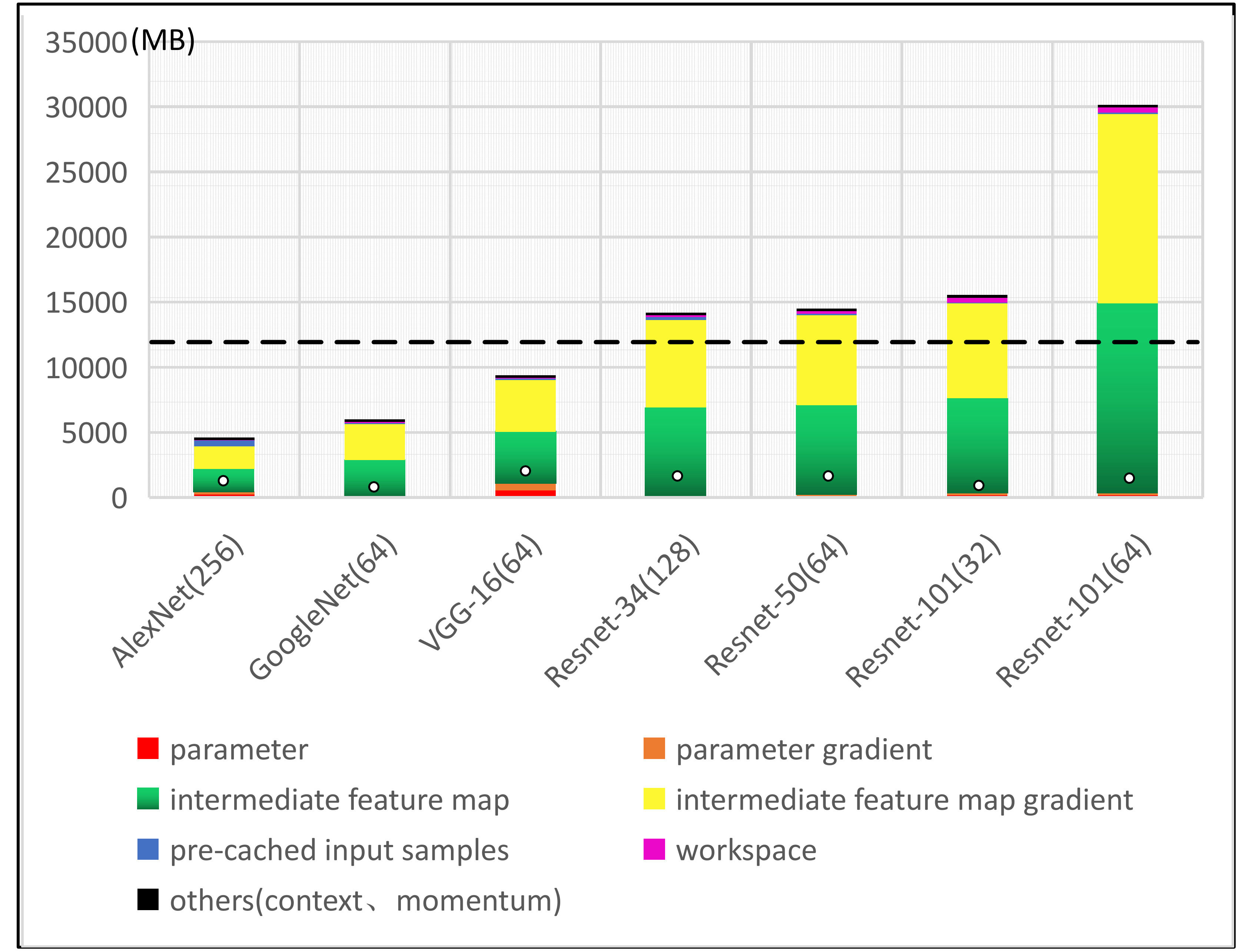}
\caption{GPU memory footprint}
\label{fig:GPU memory footprint}
\end{figure}

The data consuming memory of UDNN can be mainly categorized as: model parameters and their gradients, feature maps and their gradients(collectively called intermediate states), pre-cached input samples, necessary workspace, etc. Feature maps are the intermediate output of each layer generated in the forward propagation and reused for gradients calculation during the backward propagation. While the memory footprint of model parameters and their gradients are merely related to the network depth, that of intermediate states' also increases proportionally with the minibatch size. In general, those intermediate states occupy the vast majority of total memory footprint, as shown in Figure~\ref{fig:GPU memory footprint}.

Figure~\ref{fig:GPU memory footprint} exhibits the network-wide GPU memory footprint of several well-known CNN architectures with different minibatch size. There are two observations: 1) out-of-memory, even with a relatively small minibatch size, memory footprint of ResNet-series UDNNs still go beyond the capacity of most of mid-to-high end GPU with 12GB memory (the black dash line), as ResNet-101 with minibatch 32; 2) low GPU memory utilization, the sequential layer-wise computation only calls a small fraction of data at any moment, the white circles here denote the peak real-time memory usage, for ResNet-101 with minibatch 64, it is about 4\%, which means data more than 96\% reside in memory but don't participate in the current computation. Those prevailing deep learning frameworks, Caffe\cite{jia2014caffe}, Tensrflow\cite{abadi2016tensorflow} and PyTorch etc. haven't provided a mature solution to break the impasse.

Thus, DL practitioners only have two available alternatives to train UDNN at present. One is to decrease the minibatch size and the another is model parallelism among multiple GPUs. Indeed, these two approaches are able to break down the trainability restriction, but the adverse effects of underutilization of GPU computing resources and heavy communication overhead damage the training efficiency dramatically. However, in practical use, other than the  trainability, we should further consider the training efficiency, which is crucial to make UDNN truly applicable. For example, \cite{you2017100} points out that finishing 100-epoch ImageNet-1k training with ResNet-50 on a single NVIDIA M40 GPU takes 14 days. Such a long training time destroys the interactivity and limits productivity severely.

Now, distributed training has become the dominant scheme to accelerate the training process and the data parallelism mode is widely adopted\cite{awan2017s}. When employing distributed data parallelism mode to train UDNN, if the memory-restricted minibatch size of UDNN on single GPU gets very small, more frequent parameter synchronization among cluster would be required to finish a fixed number of epochs. As a consequence, the computation-to-communication(comp-to-comm) ratio decreases, so does the scaling efficiency. In extreme case, no speed-up achieved by distributed cluster is not impossible.


To address both the challenge in trainability and training efficiency of UDNN, this paper presents AccUDNN accelerator. Specifically, AccUDNN incorporates two core modules: memory optimizer and hyperparameter tuner. Firstly, by identifying the trainability limitation of current deep learning frameworks and the non-negligible performance degradation of the existing swap out/in optimization approaches\cite{rhu2016vdnn,wang2018superneurons}, our memory optimizer develops a dynamic swap out/in strategy to break down the trainability restriction in a way without damaging the training efficiency. The memory optimizer achieves this by carefully orchestrating what to be swapped and when to swap according to the attributes of specific network architecture and hardware resources, which is captured by the performance model. Then, after applying the memory optimization technique, the runtime environment gets complicated with host memory and PCIe communication involved, it become difficult for algorithm user who is unknown of underlying system implementation to explore the efficiency-optimal hyperparameter. Thus, the hyperparameter tuner is further designed to fix this dilemma. In a word, AccUDNN's final goal is to provide a complete and efficient scheme to train UDNN, promoting the practical use of its application.

\textbf{Our contribution}. Compared with the existing works, AccUDNN highlights in following three aspects.

1) As for the memory optimizer, our work is the first one that proposes a performance-model guided dynamic swap out/in strategy between GPU and host memory. Through quantitative analysis, we are able to pick out the suitable data to swap and plan the swap timeline more reasonably, finally eliminating the performance degradation in prior heuristic-based works effectively at the same time of achieving trainability.

2) The hyperparameter(minibatch size, learning rate) tuner is first developed from system perspective to automatically dig the optimal training efficiency after applying the memory optimization technique. Compared with other works that end up with the elementary conclusion of saving GPU memory or increasing minibatch size, we step further to explore the most relevant returns to those we really concern, i.e., training time and final accuracy.

3) AccUDNN takes both delicate system optimization and algorithmic improvement into consideration simultaneously. It provides a complete transparent solution to train UDNN, more than relieving the consumption of GPU memory.

The rest of this paper is organized as follows. Section 2 gives an overview of AccUDNN architecture. Section 3 and 4 elaborate on the design of two core modules of AccUDNN respectively. Section 5 describes the implementation of runtime memory manager. Experimental results are shown in Section 6. Section 7 covers the related research efforts and Section 8 concludes this paper.
\begin{figure}[ht]
\centering
\includegraphics[scale=0.3]{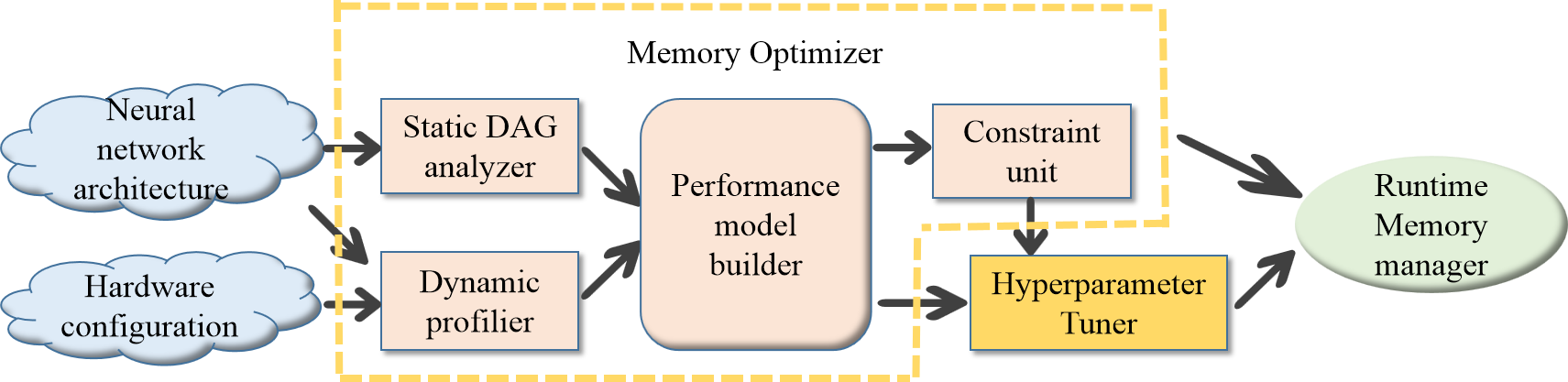}
\caption{AccUDNN architecture}
\label{fig:AccUDNN architecture}
\end{figure}

\begin{figure*}[ht]
\centering
\includegraphics[scale=0.5]{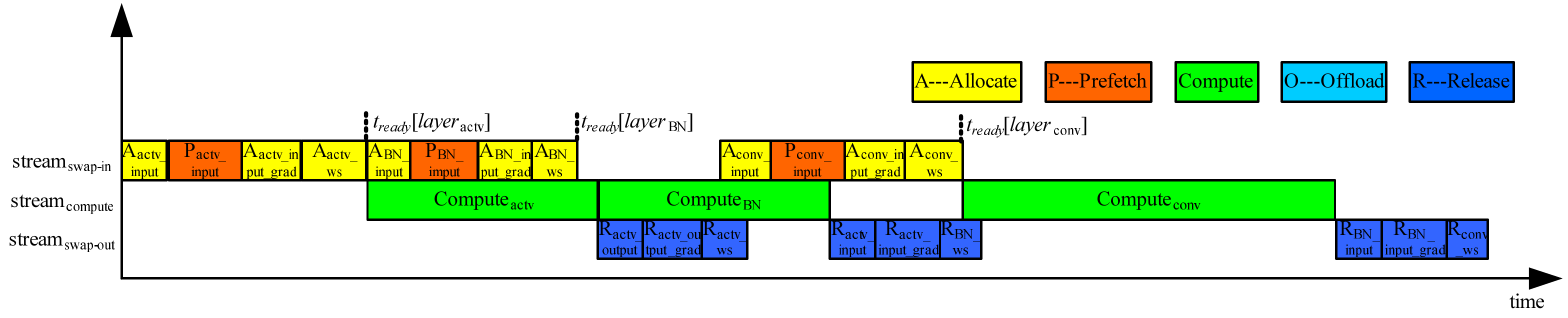}
\caption{Backpropagation with naive swap out/in operation}
\label{fig:wos_back}
\end{figure*}
\section{Overall architecture}

This section gives an overview of AccUDNN accelerator by walking through the core components, as shown in Figure~\ref{fig:AccUDNN architecture}. Given a certain UDNN and hardware configuration, the information collector first gathers both the static and dynamic attributes that are required to establish the performance model. Through data fitting, performance model builder traces the basic running behaviour of naive swap out/in mode, including GPU computational performance, GPU memory usage and PCIe communication performance. Then, the constraint unit sketches the restrictive relation among these three submodels and extracts the constraint condition that does not incur performance degradation in our dynamic swap out/in strategy. By integrating the constraint with performance model, hyperparameter tuner converts the training process after applying the memory optimization technique into an optimization problem and finally works out the efficiency-optimal hyperparameter minibatch size, then, learning rate changes accordingly. Once the optimal minibatch size is determined, the concrete swap out/in strategy(what to be swapped and when to swap) that meets the efficiency constraint gets clear as well. Finally, the concrete strategy is submitted to the runtime memory manager to deploy and execute.

\section{Memory optimizer}

Given a certain UDNN and available hardware configuration, primary thing to consider is the trainability, once trainable, the optimal training efficiency is expected. The memory optimizer develops a dynamic technique to seek the optimal training scheme. In detail, when the GPU memory is very scarce for an UDNN, it would be mandatory to swap all the involved data out to host memory to attain trainability, despite the training efficiency in this case might be unsatisfied. But usually, the GPU memory is somewhat limited(but not very scarce), it's no longer necessary to take the extreme approach. Once the trainability is ensured, we can shift the focus to the training efficiency. For training efficiency, there is a delicate balance. Swapping out as much data as possible enables larger minibatch size to be processed, which is beneficial to improve efficiency by fully utilizing GPU computing resources and reducing parameter synchronization among cluster, however, if the total volume to be transferred exceeds the PCIe bandwidth capacity, the adverse overhead would arise. Thus, memory optimizer dedicates to figure out the optimal training mode between trainability and training efficiency by dynamically adjusting the swapping strategy.


\subsection{Design principle}

For convenience, some prerequisites are given first:

1) A deep neural network with \emph{N} layers is unfolded as 2\emph{N} forward propagation processes, \emph{j} denotes layer index and $\emph{j}\in[1, 2\emph{N}]$.

2) Due to properties of high-ratio memory footprint and long time interval between two uses, the memory optimizer mainly targets feature maps.

3) Four basic operations involved in swap out/in strategy: allocate, release, offload, prefetch. Offload refers to the swap-out operation from GPU to host memory while prefetch runs in the reversed direction.

In the beginning, we select a typical convolutional unit "conv-BN-actv" as example, to give a direct view of the execution mode after incorporating the naive swap out/in strategy (swap out all the feature maps to host side). Figure~\ref{fig:wos_back} demonstrates one of the most likely scenarios. In our design, prefetch operations can continuously proceed as long as the required GPU memory is available and the interrupt will be recovered after some preceding memory occupancies be released. Because the model parameters and their gradients reside in GPU memory, the access overhead is negligible, we don't display them explicitly in Figure~\ref{fig:wos_back}. \emph{t}$_{ready}$[\emph{j}] is the timestamp that marks all the data required by the computation of layer \emph{j} have swapped in GPU memory and the necessary workspace have allocated, before this point, GPU cannot execute layer \emph{j}'s computation and may enter into stall state(refers to the scenario of GPU computation suspending and waiting for the required data transfer operations get ready). It is observed that the computation of BN layer follows activation layer compactly while an undesirable stall duration lies between BN and convolution layer. The fast computation of BN layer, associate with the memory-restricted delayed initial point results in inefficient overlapping of prefetch operation P$_{conv\_input}$ and computation. In this case, the input feature map of convolution layer is not suitable to swap out in terms of efficiency.

\begin{figure}[ht]
\centering
\includegraphics[scale=0.42]{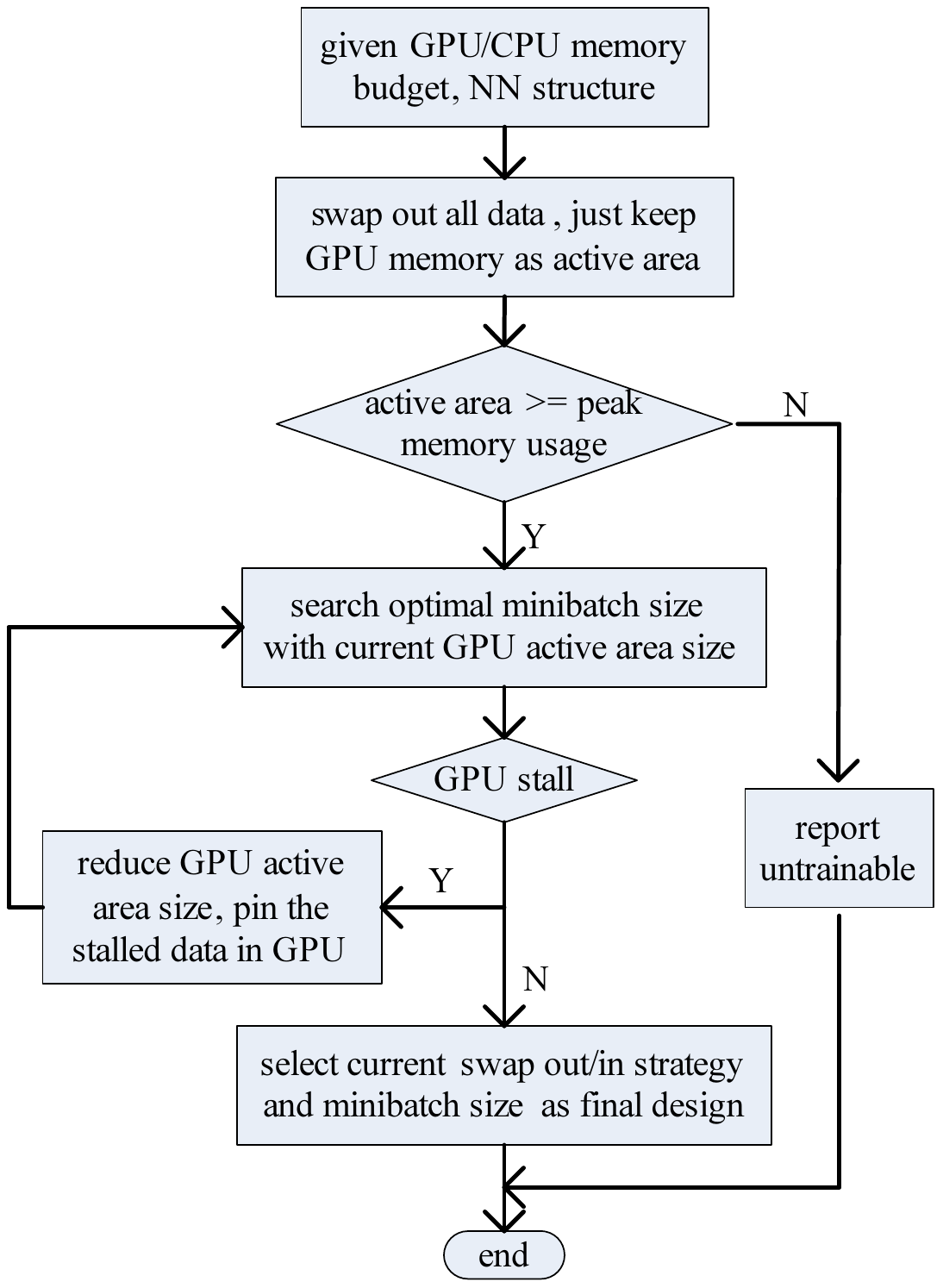}
\caption{Flow chart of dynamic swap out/in strategy}
\label{fig:flow chart}
\end{figure}

The flow chart of our dynamic swap out/in strategy is shown in Figure~\ref{fig:flow chart}. Two key steps are highlighted here.

First, to maximize the trainability, the extreme case is to swap all the feature maps out to host side, just leave GPU memory as layer-wise active area for necessary data and workspace accessed by current layer's computation. A peak layer-wise memory usage can be obtained by traversing the allocate/release operations in the global memory-object access pattern(GMAP) and is proportional to the minibatch size. As long as the GPU memory budget is larger than the peak layer-wise memory usage, the UDNN is trainable and the processible minibatch size reaches its maximum at this point. The reduction of GPU memory usage from network-wide to layer-wide removes original depth limitation and enables much larger minibatch size. In this case, the running mode can be compared to the interaction between cache and DRAM in traditional computer architecture, here, GPU memory plays the role of cache. Fortunately, the memory access pattern of deep learning task is explicitly layer-wise and can be obtained by static analysis or pre-run easily, which greatly facilitates the design of our strategy.

Then, if the trainable condition is satisfied, the memory optimizer moves forward to explore the efficiency-optimal swap out/in strategy. We employ a performance model to exam whether the data volume to be transferred with current minibatch size will stall GPU computation, if does, we gradually shrink the minibatch size, freeing up partial GPU memory from the active area to pin those stalled feature maps and don't swap them out any more. By gradually adjusting, the expected swap out/in strategy with no performance overhead can be found finally.

\subsection{Information gathering}
To build the performance model, information collector needs to capture two types of information, static and dynamic information, to profile the necessary characteristics of the given UDNN and hardware configuration.

\textbf{Static information}. For a given UDNN with user-specified minibatch size \emph{k$_{base}$}, by static analysis of the dataflow graph, we can collect:

1) global memory-object access pattern (GMAP), stacked by four basic swap out/in operations, denotes the data access and release pattern within one iteration. The GMAP is invariable across iterations and independent with minibatch size.

2) data size of every operation in GMAP, equals to memory consumption and PCIe transfer volume. As minibatch size maps to one channel of input, the corresponding size of feature maps related operation is proportional to minibatch size.

3) single precision floating-point operations (FLOPs) of per layer, similar to 2), it grows linearly with the minibatch size. Both 2) and 3) can be calculated by simple analysis of tensor dimension, besides, the results under arbitrary minibatch sizes can be figured out easily with record of base minibatch size and the proportional relation instead of starting from scratch.

Static information profiling phase can be executed by simple graph traverse and basic arithmetical operation without actual running the training process.

\textbf{Dynamic information}. We adopt a costless profiling stage to gather dynamic information, in detail, we set minibatch size as 1/8, 1/4, 1/2, 2/3 and 1 of the maximal trainable minibatch size respectively in the first five epochs and employ the naive swap out/in strategy without regarding to the stall problem. In the runtime of these five epochs, we collect:

1) per layer's computation time in one iteration of different minibatch sizes.

2) per swap operation's PCIe transfer time in one iteration of different minibatch sizes.

The dynamic information collection is conducted by \emph{nvprof}, a handy profiling tool provided by Nvidia. The profiling approach benefits in three aspects. First, it can model various hardware configurations(GPU architecture, PCIe bandwidth) automatically without any user intervention. Furthermore, the results are much closer to practical condition compared with the nominal values, which profits the effective design of our swapping strategy. Lastly, integrated as early epochs of the whole training process, the profiling stage is still contributory.
\subsection{Performance model}

The first key point in the memory optimizer is to identify whether and when GPU computation stalls. We develop a performance model to handle this. What's more, with the performance model, we are able to design the swap out/in strategy in a more fine-grained manner, effectively eliminating the system performance degradation of heuristic design. The performance model includes three parts: GPU computation model, GPU memory usage model, PCIe communication model.

\subsubsection{GPU computational model}

\begin{figure}[ht]

\centering
\includegraphics[scale=0.12]{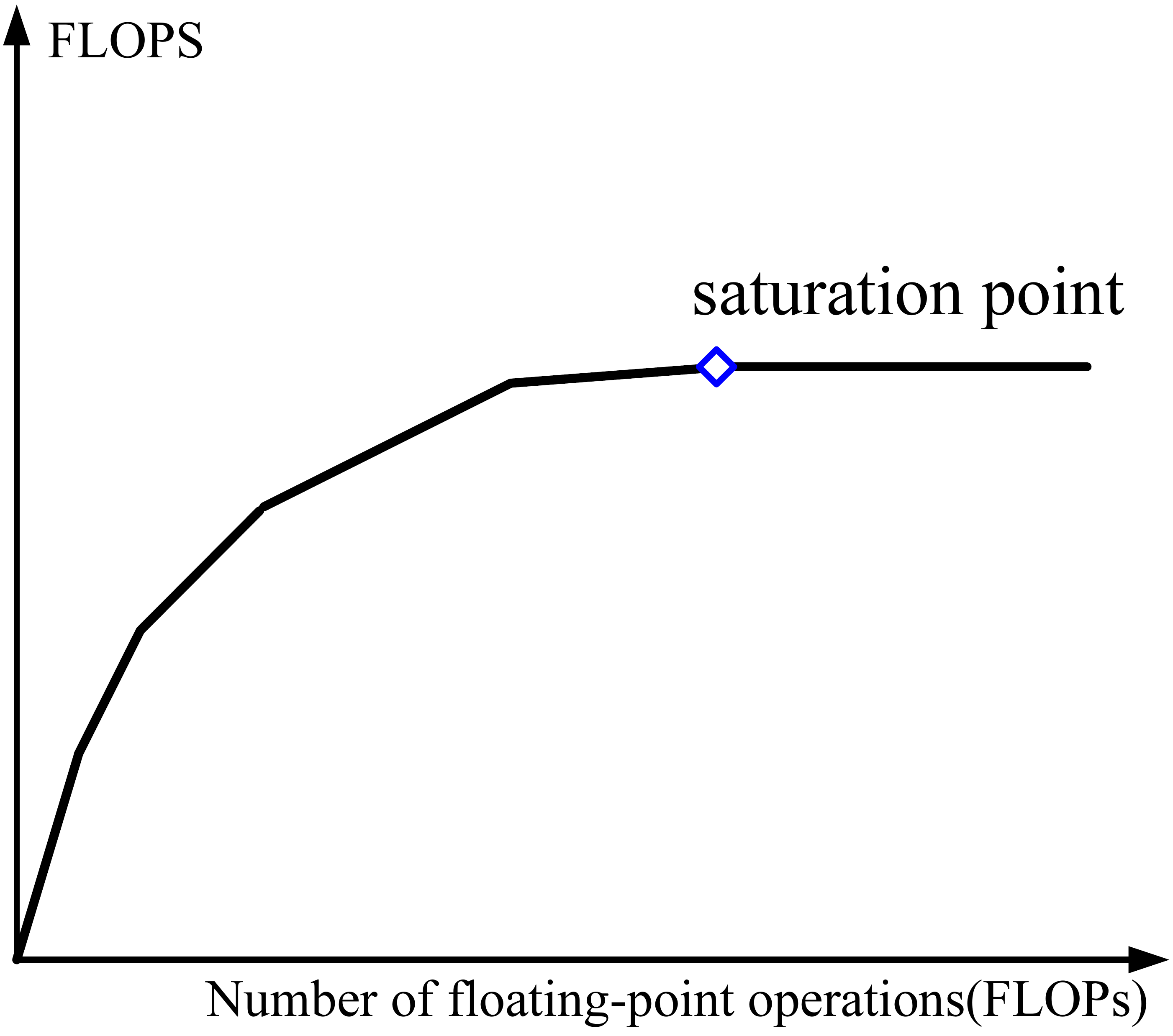}
\caption{Trend of GPU computational performance}
\label{fig:GPU computational efficiency schematic}
\end{figure}

GPU computational performance for given UDNN is first modeled. All the collected information tuples, per layer's FLOPs and computation time, among five iterations with different minibatch sizes are categorized to multiple groups by layer type(conv, BN, pooling, FC, etc.). Then, the computing performance for each layer type is modeled respectively. Via fitting, we can get the general performance trend of all the layer types, displayed in the schematic diagram Figure~\ref{fig:GPU computational efficiency schematic}. From the performance curve, we have two observations. One is that the GPU throughput FLOPS grows gradually with a certain range owing to the trend of more parallel computing resource (processor units, memory bandwidth) putting into use. The another one is that there is a performance saturation point(the blue rhombus) and the throughput remains unchanged when a FLOPs exceeds the value of threshold point, that means, the execution of a kernel with extreme large FLOPs will be converted to a serial steps and each step uses up all the computing resources. The saturation point identifies the maximum available degree of parallelism, besides, the FLOPS at saturation point of each layer type may differs from the others', which depends on their concrete implementation.

By querying the corresponding FLOPS from GPU computational performance model, it is handy to calculate a layer's computation time under arbitrary minibatch size \emph{k}. Then, the execution time of one iteration and whole training process can be modelled successively.

Formally, given a \emph{N}-layer UDNN, layer index \emph{j}, minibatch size \emph{k}, a constant number of training epochs \emph{E}, dataset size \emph{m}. Per layer's floating-point operation is defined as \emph{FLOPs}$_{\emph{j}}$(\emph{k}). Based on static information 3), we have

\begin{equation}
FLO{{Ps}_{^{j}}}(k)=\frac{k}{{{k}_{base}}}FLO{{Ps}_{^{j}}}({{k}_{base}}),j\in [1\text{, }2N]
\end{equation}

With per layer's computation time \emph{t$_{j}$}(\emph{k}), execution time for one iteration \emph{t$_{iter}$}(\emph{k}) approximately equals to the sum of \emph{t$_{j}$}(\emph{k}) due to the elimination of GPU computation stall in our swapping strategy, thus

\begin{equation}
\begin{split}
  &{{t}_{iter}}(k)=\sum\limits_{j=1}^{2N}{{{t}_{j}}(k)}=\sum\limits_{j=1}^{2N}{\frac{FLO{{Ps}_{j}}(k)}{FLOPS(FLO{{Ps}_{j}}(k))}} \\
 &=\sum\limits_{j=1}^{2N}{\frac{\frac{k}{{{k}_{base}}}FLO{{Ps}_{j}}({{k}_{base}})}{FLOPS(\frac{k}{{{k}_{base}}}FLO{{Ps}_{j}}({{k}_{base}}))}}
\end{split}
\end{equation}
\setlength\parindent{0em}and execution time for whole training process \emph{t$_{whole}$}(\emph{k}) is

\begin{equation}
\begin{split}
  & {{t}_{whole}}(k)=\frac{E*m}{k}\left[ {{t}_{iter}}(k)+\Delta  \right] \\
  &=\frac{E*m}{k}\left[\sum\limits_{j=1}^{2N}{\frac{\frac{k}{{{k}_{base}}}FLO{{Ps}_{j}}(base)}{FLOPS(\frac{k}{{{k}_{base}}}FLO{{Ps}_{j}}(base))}}+\Delta  \right] \\
 & =E*m\sum\limits_{j=1}^{2N}{\left[ \frac{\frac{1}{{{k}_{base}}}FLO{{Ps}_{j}}({{k}_{base}})}{FLOPS(\frac{k}{{{k}_{base}}}FLO{{Ps}_{j}}({{k}_{base}}))} \right]}+\frac{E*m}{k}\Delta
\end{split}
\end{equation}
where $\Delta$ denotes the time expenditure between two iterations, in single GPU mode, it is negligible with pre-cached input samples, however, additional time for parameter synchronization among cluster in distributed mode must be taken into account. Noting that there are two parts in \emph{t$_{whole}$}(\emph{k}). The first part is pure GPU computation time, with the increase of minibatch size \emph{k}, denominator FLOPS(*) also increases until it reaches the threshold point, thus, the computation time gets shorter and levels off finally. Value of minibatch size \emph{k} at threshold point hinges on specific deep neural network and hardware resource. The second part makes sense chiefly in distributed mode, especially for those networks with massive model parameters, reducing times for parameter synchronization by increasing minibath size \emph{k} is a most straightforward and effective method.

\subsubsection{GPU memory usage model}
We take two independent streams, stream$_{swap-out}$ and stream$_{swap-in}$, to execute the offload and prefetch operations in background respectively. In our design, an active area with size of peak memory usage is assigned in GPU memory to serve cyclically for per layer's computation. In the layer-wise propagation, peak memory usage can be achieved within finite times even only once, thus, the active area is underutilized in most cases. This fact provides stream$_{swap-in}$ with opportunity to swap in subsequent data of other layers as soon as possible, which is helpful to ease GPU computation stall problem. With allocate operation consuming GPU memory and release operation freeing memory concurrently, the real-time memory usage must not be greater than the given GPU memory budget \emph{M$_{budget}$}. Then, at any time, the finished subsequences \emph{Seq$_{allocate}$} and \emph{Seq$_{release}$} satisfy constraint (4), in which data size of each operation \emph{M$_{allocate/release}$}(\emph{k}) is obtained with static information 2).
\begin{equation}
\begin{split}
   & \sum\limits_{p=0}^{Se{{q}_{allocate}}}{\frac{k}{{{k}_{base}}}{{M}_{allocate[p]}}({{k}_{base}})}- \\
    &\sum\limits_{q=0}^{Se{{q}_{release}}}{\frac{k}{{{k}_{base}}}{{M}_{release[q]}}({{k}_{base}})}\le{{M}_{budget}}
\end{split}
\end{equation}

\subsubsection{PCIe communication model}
PCIe gen3 provides a maximum data transfer bandwidth of 16 GB/s between host and GPU device. By averaging a group of actual transfer rates that derived from static information 2) and dynamic information 2), a more down-to-earth PCIe bandwidth \emph{Bandwidth$_{avail}$} in current hardware environment is obtained. Because the feature maps are generated in forward propagation and reused in backward propagation to calculate the parameter gradients, as a result, offload operation mainly happens in forward process while prefetch operation happens inversely in backward process. The runtime of prefetch/offload operation under minibatch size \emph{k} is formulated as (5) where \emph{T$_{pre/off}$}(*) represents the data size to be transferred. If a feature map is pinned in GPU memory, \emph{t$_{pre/off}$} = 0.
\begin{equation}
\begin{split}
 &{{t}_{pre/off}}={\frac{\frac{k}{{{k}_{base}}}{{T}_{pre/off}}({{k}_{base}})}{Bandwidt{{h}_{avail}}}}.
\end{split}
\end{equation}

\subsection{Constraint condition}
The above three submodels are interrelated and mutually restricted, impacting the performance of swap out/in strategy together. To avoid stalling GPU computation, we define a time sequence \emph{t$_{ready}$} with width of 2\emph{N} in the range of a single iteration, in which each element identifies a timestamp after when layer \emph{j} can start computing at any moment. That means, all the required data have been swapped in GPU memory and necessary workspace have been allocated, layer \emph{j}'s kernel function is ready to be called by stream$_{compute}$. Algorithm 1 demonstrates the key steps of determining \emph{t$_{ready}$}.
\vskip 0.31cm
\par\noindent\rule[1pt]{8.5cm}{0.08em}
\par\setlength\parindent{0em}\textbf{Algorithm 1}: Determining the time sequence \emph{t$_{ready}$}
\par\noindent\rule[1pt]{8.5cm}{0.08em}
\par\noindent\textbf{Input}: subsequences \emph{Seq$_{allocate}$} and \emph{Seq$_{release}$}, GPU memory budget \emph{M$_{budget}$}
\par\noindent\textbf{Output}: time sequence \emph{t$_{ready}$}
\par\noindent\emph{t$_{ready}$}[\emph{j}] = {0}, \emph{j }is the layer index;
\par\setlength\parindent{0em}\emph{M$_{used\_present}$} = 0;
\par\setlength\parindent{0em}for (\emph{j} = 1; \emph{j} $\leq$ 2\emph{N}; \emph{j}++) \{
\par\setlength\parindent{1em}if (\emph{j} == 1) \{
\par\setlength\parindent{2em}\emph{t$_{ready}$}[\emph{j}] = $\sum${$_{\emph{j}}$}\emph{t$_{pre}$} sum of \emph{t$_{pre}$} for layer \emph{j}'s prefetch
\par\setlength\parindent{2em}operations;
\par\setlength\parindent{1em}\} else if (\emph{M$_{used\_present}$} +
{size of \emph{Seq}}$_\emph{allocate}^\emph{j}$ $\leq$ \emph{M$_{budget}$})\{
\par\setlength\parindent{2em}\emph{t$_{ready}$}[\emph{j}] = \emph{t$_{ready}$}[\emph{j}-1] + $\sum${$_{\emph{j}}$}\emph{t$_{pre}$};
\par\setlength\parindent{1em}\} else \{
\par\setlength\parindent{2em}Estimate current layer \emph{j}$_{\emph{current}}$ be executing by
\par\setlength\parindent{2em}comparing \emph{t$_{ready}$}[\emph{j}-1] and $\sum${$_1^\emph{current}$}\emph{t}$_{\emph{compute}}$;
\par\setlength\parindent{2em}Traverse Seq$_{release}$ from layer \emph{j}$_{\emph{current}}$ to \emph{j}$_{\emph{satisfied}}$
\par\setlength\parindent{2em}until find \emph{j}$_{\emph{satisfied}}$ that satisfies sum of the released
\par\setlength\parindent{2em}memory from layer \emph{j}$_{\emph{current}}$ to \emph{j}$_{\emph{satisfied}}$ larger than
\par\setlength\parindent{2em}{size of \emph{Seq}}$_\emph{allocate}^\emph{j}$;
\par\setlength\parindent{2em}\emph{t$_{ready}$}[\emph{j}] = $\sum${$_1^\emph{satisfied}$}\emph{t}$_{\emph{compute}}$ + $\sum${$_{\emph{j}}$}\emph{t$_{pre}$};
\par\setlength\parindent{1em}\}
\par\setlength\parindent{1em}\emph{M$_{used\_present}$} += size of {\emph{Seq}}$_\emph{allocate}^\emph{j}$;
\par\setlength\parindent{0em}\}
\par\setlength\parindent{0em}return \emph{t$_{ready}$};
\par\noindent\rule[1pt]{8.5cm}{0.08em}
\vskip 0.31cm

Additional remarks for Algorithm 1. Because the time cost for memory allocation operation is trivial, we primarily consider the prefetch operation here that has a direct bearing on the proceeding of computation. On condition of trainable, the involved data of first layer can be prefetched directly at the beginning of iteration. If GPU memory is sufficient, prefetch operation of subsequent layers run continuously without interruption, \emph{t$_{ready}$}[\emph{j}] equals to the accumulation of communication time. Once no memory available, prefetch operation will be suspended to wait for other memory space to release. According to the interrupted timestamp, we can estimate layer \emph{j}$_{\emph{current}}$ that stream$_{compute}$ is executing now, then, by traversing Seq$_{release}$ from \emph{j}$_{\emph{current}}$ to \emph{j}$_{\emph{satisfied}}$ until memory to be released in this range satisfies the requirement of suspended layer, \emph{j}$_{\emph{satisfied}}$ is picked out. Prefetch operation of the suspended layer can't be executed until layer \emph{j}$_{\emph{satisfied}}$'s computation finished, so, the corresponding \emph{t$_{ready}$}[\emph{j}] equals to the sum of computation time from layer 1 to \emph{j}$_{\emph{satisfied}}$ and the prefetch communication time. Notice that real-time memory usage \emph{M}$_{\emph{used\_present}}$ always increases in Algorithm 1, actually, with the proceeding of computation, memory release operation conducted by stream$_{swap-out}$ also decrease \emph{M}$_{\emph{used\_present}}$, we just don't depict it explicitly here.
With \emph{t$_{ready}$}, the constraint of without stalling GPU computation is expressed as

\begin{equation}
{{t}_{ready}}[j]<=\sum\limits_{r=1}^{j-1}{t_{r}^{{}}}
\end{equation}
\par\setlength\parindent{0em}the ready timestamp of layer \emph{j} must be ahead of the accomplishment of layer (\emph{j}-1)'s computation.

\section{Hyperparameter tuner}
This section elaborates on solving the efficiency-optimal minibatch size and matched learning rate in the new runtime environment with host memory and PCIe communication involved after applying the memory optimization technique. By integrating the training time and constraints of swapping strategy as a general optimization problem, the efficiency-optimal minibatch size and its corresponding swap strategy are jointly solved. Besides, to guarantee the final accuracy, we figure out an adaptive rule of the learning rate to accommodate the efficiency-optimal minibatch size.
\subsection{Efficiency-optimal minibatch size}
In the light of performance model and the constraint, the whole training process with swap out/in strategy can be formulated as an optimization problem (7) with respect to minibatch size
\begin{equation}
\begin{split}
 & min \quad (3) \\
 & s.t. \quad (4)\ \&\ (6)
\end{split}
\end{equation}
Equation (7) can be categorized as integer linear programming problem, we build a linear search algorithm to determine the efficiency-optimal minibatch size. See Algorithm 2. The search process starts with the maximum minibatch size and drops to the efficiency-optimal point where the two constraints are both satisfied. At the same time, which feature maps to be swapped or pinned under the efficiency-optimal minibatch size is also identified.
\vskip 0.31cm
\par\noindent\rule[1pt]{8.5cm}{0.08em}
\par\setlength\parindent{0em}\textbf{Algorithm 2}: Determining the efficiency-optimal minibatch size
\par\noindent\rule[1pt]{8.5cm}{0.08em}
\par\setlength\parindent{0em}\textbf{Input}: GMAP, GPU memory budget \emph{M$_{budget}$}
\par\setlength\parindent{0em}\textbf{Output}: Efficiency-optimal minibatch size \emph{k}$^*$ and its corresponding swapping strategy
\par\setlength\parindent{0em}Traverse GMAP to get the Seq$_{peak}$ when achieve peak memory usage;
\par\setlength\parindent{0em}\emph{k}$_{\emph{max}}$ = \emph{k}$_{\emph{base}}$*[\emph{M$_{budget}$} - \emph{M$_{others}$} - \emph{M$_{para}$}(in Seq$_{peak}$) - \emph{M$_{ws}$}(in Seq$_{peak}$)] / {$\sum$[ \emph{M$_{FM}$}(\emph{k}$_{\emph{base}}$)(in Seq$_{peak}$)]};
\par\setlength\parindent{0em}\emph{k} = \emph{k}$_{\emph{max}}$;
\par\setlength\parindent{0em}Calculate \emph{t$_{ready}$} under \emph{k} with constraint (4);
\par\setlength\parindent{0em}while (constraint (6) is not satisfied under \emph{k} for all layer 1 to 2\emph{N}) \{
\par\setlength\parindent{1em}\emph{k} -= 1;
\par\setlength\parindent{1em}Calculate \emph{t$_{ready}$} under \emph{k} with \emph{M$_{budget}$} = \emph{M$_{peak}$}(\emph{k})
\par\setlength\parindent{1em}in constraint (4);
\par\setlength\parindent{1em}Identify layer index set ${\Omega}$ that don't satisfy (6);
\par\setlength\parindent{1em}Try to pin data involved in ${\Omega}$ in GPU memory area
\par\setlength\parindent{1em}(\emph{M$_{budget}$} - \emph{M$_{peak}$}(\emph{k}));
\par\setlength\parindent{0em}\}
\par\setlength\parindent{0em}\emph{k}$^*$ = \emph{k};
\par\setlength\parindent{0em}return the efficiency-optimal minibatch size \emph{k}$^*$;
\par\noindent\rule[1pt]{8.5cm}{0.08em}
\vskip 0.31cm

\subsection{Adaptive learning rate}
\par\setlength\parindent{0em}After determining the efficiency-optimal minibatch size from a systematic perspective, the adaptive learning rate should also be taken into account as algorithmic improvement. Only by the co-design of system and algorithm can we achieve both satisfied training time and accuracy. From practice experience, with a constant number of epochs, larger minibatch size usually leads to final accuracy loss if maintain the same learning rate with those small minibatch sizes. Hence, we try to overcome this barrier by borrowing the idea in \cite{bottou2018optimization} to adjust learning rate adaptively. Theorem 4.6 in \cite{bottou2018optimization} demonstrates the convergence property with fixed learning rate. To retain the same convergence rate under different minibatch size, the contraction item should meet

\begin{equation}
{{(1\text{-}{{\alpha }_{base}}c\mu )}^{iter{{s}_{base}}-1}}={{(1\text{-}{{\alpha }^{*}}c\mu )}^{\frac{iter{{s}_{base}}}{q}-1}}
\end{equation}
\par\setlength\parindent{1em}In (8), $\alpha$$_{\emph{base}}$ and \emph{iters}$_{\emph{base}}$ denote learning rate and iterations under minibatch size \emph{k}$_{\emph{base}}$ while $\alpha$$^*$ is the learning rate to be solved that matches with \emph{k}$^*$. \emph{q} is multiple between \emph{k}$^*$ and \emph{k}$_{\emph{base}}$. The gradient of large minibatch size is relatively close to unbiased estimate, then $\mu$ $\approx$ 1. \emph{c} represents convexity of objective function. By solving (8),

\begin{equation}
 {{\alpha }^{\text{*}}}\approx \frac{1-{{(1-{{\alpha }_{base}}c)}^{q}}}{c}\
\end{equation}

\par\setlength\parindent{1em}With larger minibatch size, the reduction of variance of stochastic gradient direction allows a bigger step in each iteration, compensating for deficiency of fewer parameter update times in an alternative way.

%
%
%
%

\section{Runtime Memory Manager}
The runtime memory manager provides four basic memory access operations: allocate, release, offload, prefetch. Allocate/release operation includes memory allocation and release in both GPU and host side. In GPU side, we first allocate all the given memory budget as an integrated buffer pool with one \emph{cudaMalloc}() call when a new iteration starts, the memory manager then allocates GPU buffer from the buffer pool directly for each allocate operation at runtime, accordingly, the release operation returns the buffer space back to the buffer pool. The memory space in host side is allocated in pinned way, removing the time cost of the copy step between pageable and pinned memory area. Offload/prefetch operation copies the required data between GPU and host memory by PCIe communication.

To overlap the GPU computation and data transfers, The runtime memory manager deploys three separate CUDA streams for computation, swap-out and swap-in. Such a design is well-matched to most of GPU architectures with a kernel engine and two copy engines(Device-to-Host copy engine and Host-to-Device copy engine). stream$_{swap-in}$ manages the allocate and prefetch operations while stream$_{swap-out}$ manages the release and offload operations. For swap-out operation, stream$_{swap-out}$ waits for notification from stream$_{compute}$, once a feature map generated, stream$_{swap-out}$ starts offload operation immediately. For swap-in operation, stream$_{swap-in}$ prefetches successively until no GPU memory is available and the interrupt will be recovered after stream$_{swap-out}$ releases some preceding memory occupancies. The offload operation mostly takes place in the forward propagation phase and the prefetch operation corresponding to the backward propagation phase. Offload as soon as possible is beneficial to spare more memory space for posterior executions, likewise, prefetch early can avert stalling GPU computation as far as possible.

\section{Experiment}

\subsection{Experimental setup}
AccUDNN is evaluated on those representative competitors in ILSVRC12 classification task, particularly the UDNN ResNet and its variants. All the models are trained for 90 epochs. Each server has one NVIDIA M40 GPU with 24GB GPU device memory and 128GB RAM. Eight servers interconnected with 10Gbps Ethernet constitute the distributed cluster. We implement AccUDNN on the open-source framework Caffe with slight modification. The overall training procedure is split into three stages. We begin with the profiling stage in first five epochs with user-specified learning rate, then, a decision-making process is conducted with the established performance model to single out the efficiency-optimal minibatch size, corresponding swapping strategy and the matched learning rate, finally, run the remaining epochs with the optimal configuration till finish.

\subsection{Verification of performance model}
We pick AlexNet as example to verify the correctness of performance model.

\subsubsection{GPU computational model}
\begin{figure}[ht]
\centering
\includegraphics[scale=0.2]{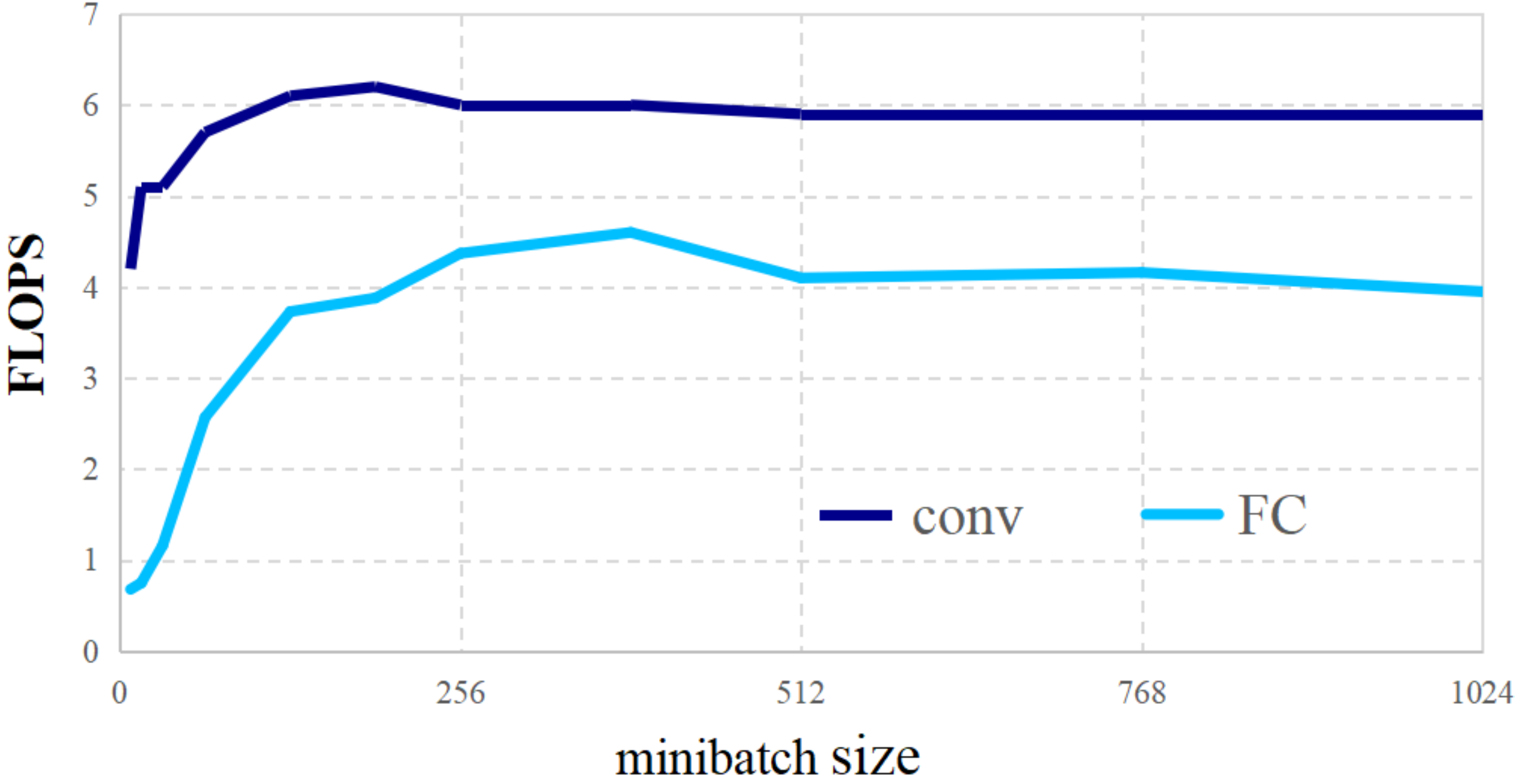}
\caption{Actual GPU computational performance}
\label{fig:actual GPU computational efficiency}
\end{figure}

\textbf{\emph{A}. GPU computational efficiency}. Figure~\ref{fig:GPU computational efficiency schematic} is a schematic diagram that presents the general trend of GPU computational performance. Here, take major time-consuming layer types, conv and FC, as representatives, actual computational performance with cuDNN implementation is demonstrated as Figure~\ref{fig:actual GPU computational efficiency}. Along with the increase of minibatch size from 8 to 1024, the increase of actual FLOPS gradually slows down to a plateau, which is consistent with the conclusion of Figure~\ref{fig:GPU computational efficiency schematic}. The exact FLOPS of a certain minibatch size is gained by averaging over all the layers of same type within one iteration.

\begin{figure}[ht]
\centering
\includegraphics[scale=0.2]{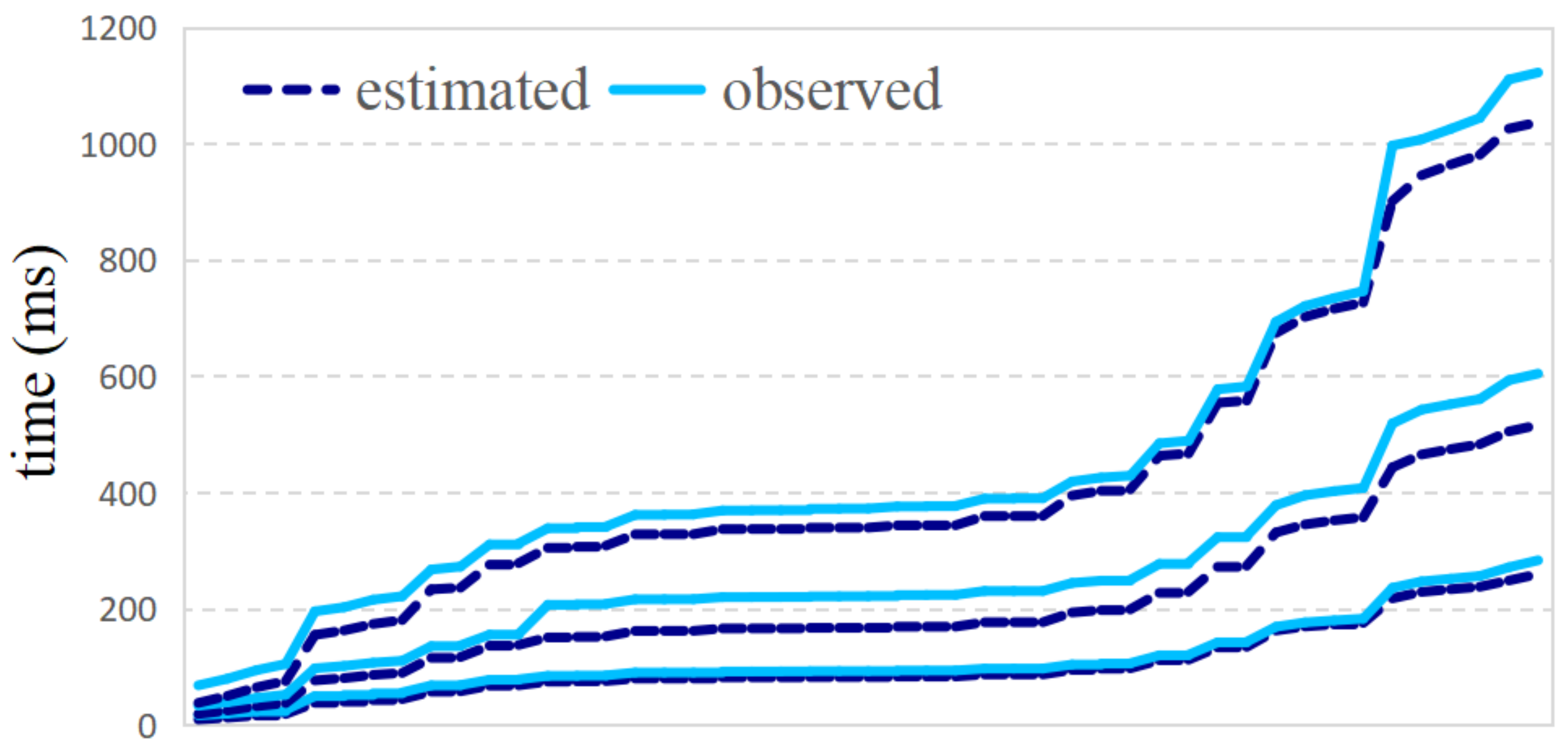}
\caption{Estimated and observed training timeline}
\label{fig:training_time}
\end{figure}

\textbf{\emph{B}. Training time}. With formula (2) and Figure~\ref{fig:actual GPU computational efficiency}, we first estimate the layer-wise execution time successively and then get the training timeline within one iteration. The comparisons between the estimated value and observed value of minibatch size 256, 512 and 1024 are exhibited in Figure~\ref{fig:training_time}. It's observed that there is a relative small error the estimated curve lags, which is caused by the unmodeled overhead of schedule and startup of kernel. To estimate more precisely, we reduce the denominator FLOPS(*) in (2) slightly to take the unmodeled overhead into account in practical use.

\subsubsection{GPU memory usage model}
\begin{figure}[ht]
\centering
\includegraphics[scale=0.22]{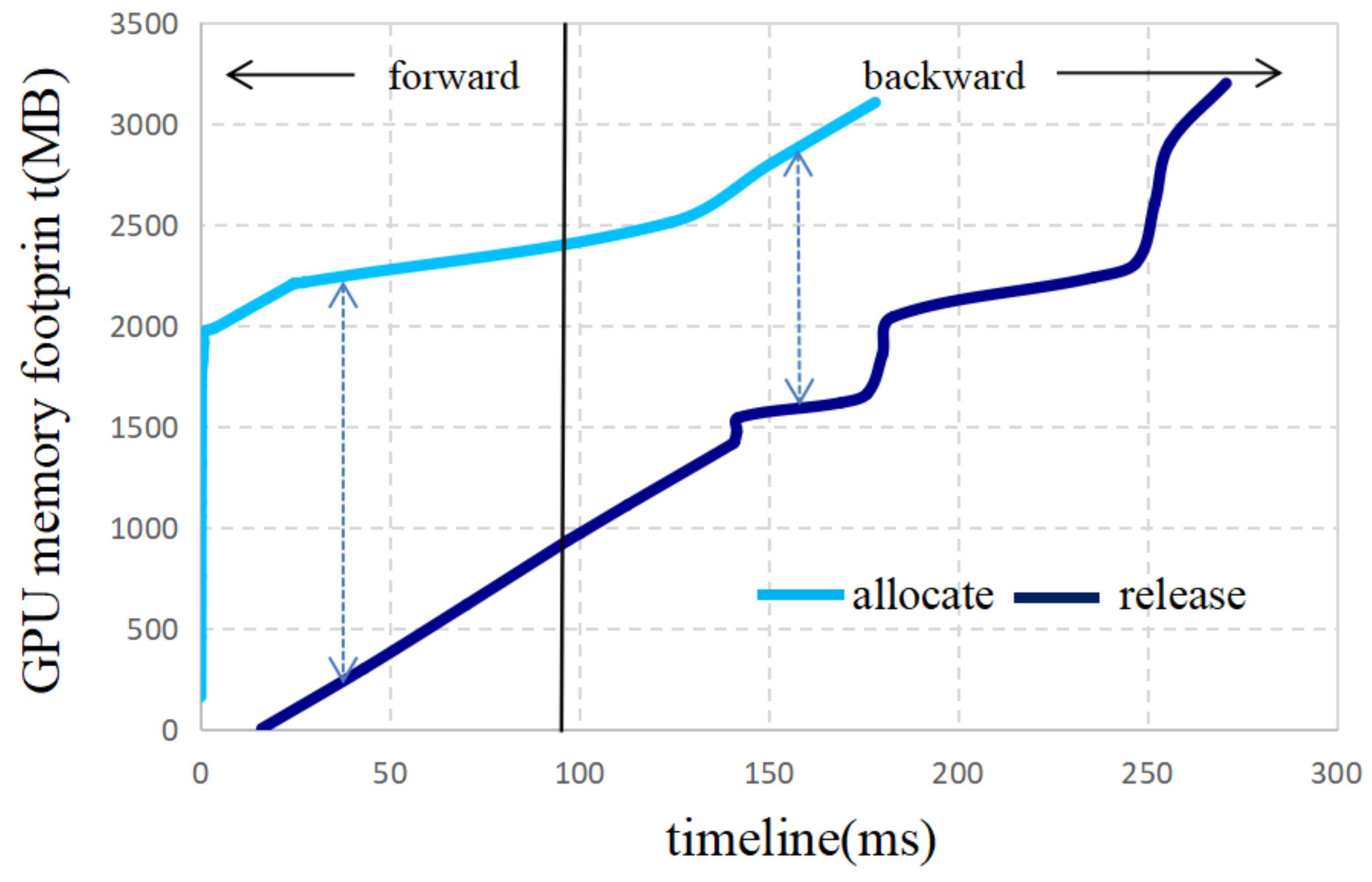}
\caption{GPU memory usage within one iteration}
\label{fig:memory usage}
\end{figure}

When adopting AccUDNN to train AlexNet(256) within 3GB GPU memory budget, the GPU memory usage of stream$_{swap-in}$ and stream$_{swap-out}$ in one iteration is shown as Figure~\ref{fig:memory usage}. When a new iteration starts, stream$_{swap-in}$ invokes the allocate operation to allocate GPU memory for feature maps and workspace in forward propagation continuously, because the time cost of allocate operation is usually in the level of microseconds, massive memory space gets occupied rapidly (the straight up area). stream$_{compute}$ switches to backward propagation around 100ms. In backward propagation process, stream$_{swap-in}$ also invokes prefetch operation to swap in the feature maps, the speed of memory occupation slows down visibly due to the data transfer. stream$_{swap-out}$ always follows stream$_{compute}$, invoking offload operation to swap out the feature maps in forward propagation process and release operation to free up memory space in backward propagation process. stream$_{swap-out}$ finally frees up all the memory allocated by stream$_{swap-in}$, but at any moment, the different between two curves (the dashed two-way arrow) must be less than the given memory budget, this is what described in formula (4).

\subsubsection{PCIe communication model}
The PCIe communication model is very intuitive, we won't go into details here.

\begin{figure}[ht]
\centering
\includegraphics[scale=0.36]{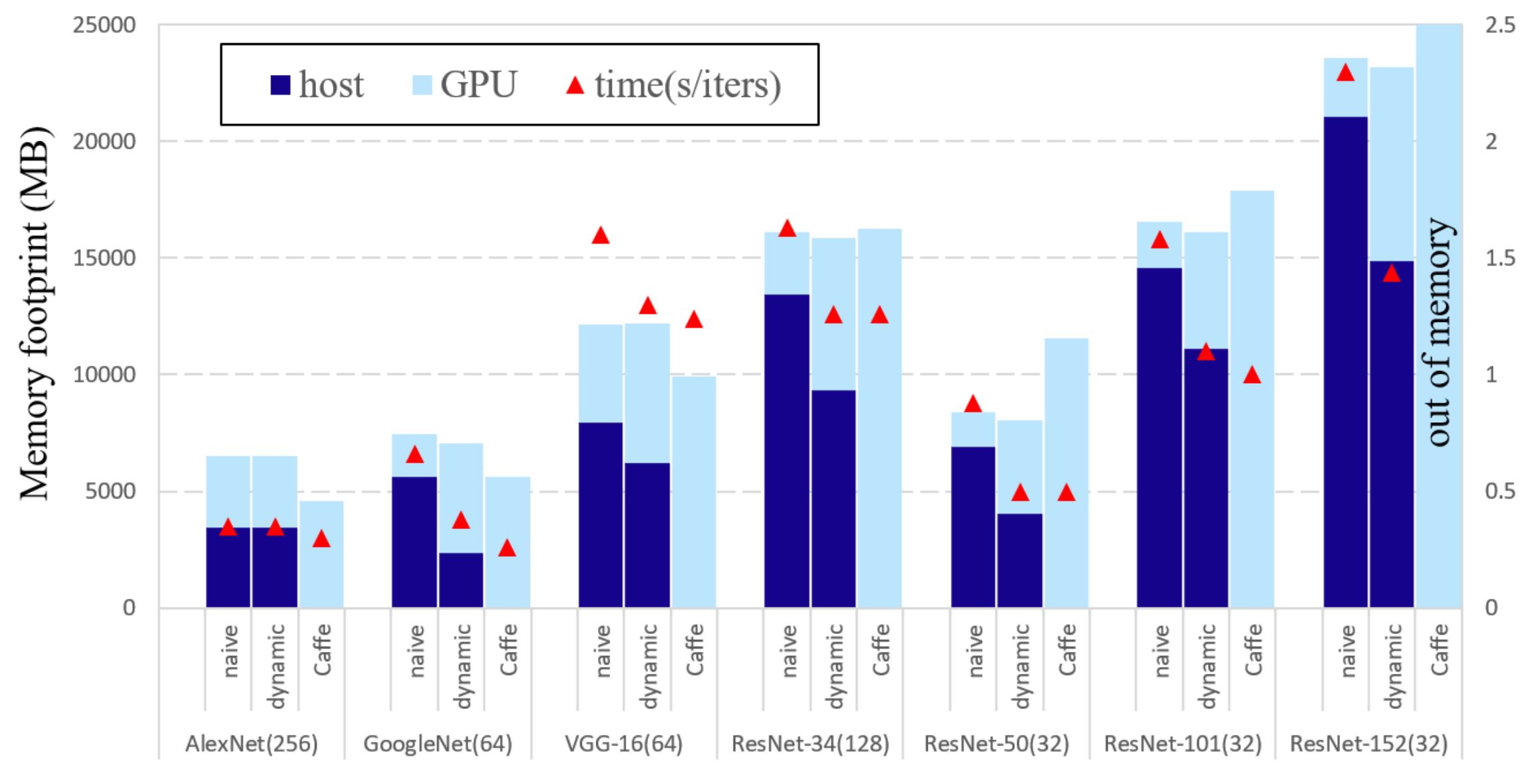}
\caption{Evaluation of dynamic swap out/in strategy}
\label{fig:memory reduction}
\end{figure}

\subsection{Evaluations of AccUDNN}
\textbf{\emph{A}. Dynamic swap out/in strategy}. Figure~\ref{fig:memory reduction} presents the memory footprint for seven well-known UDNNs. Three comparative cases are studied here. The "naive" refers to the naive swap out/in strategy crudely offload all the feature maps to host memory and execute without regard to the non-stall constraint. The "dynamic" refers to our dynamic swapping strategy proposed in the memory optimizer. The "Caffe" refers to the original Caffe that keeps all the data in GPU.

It shows that compared with the original Caffe, naive swap out/in strategy greatly reduces the trainable GPU memory requirement. Typically, for the UDNN ResNet-152 with minibatch size 32, from more than 24GB to 2.5GB, the swap out/in strategy shrinks the GPU memory requirement from network-wide level to layer-wide level successfully. However, the minimum trainable GPU memory capacity stalls the computation to different degrees which leads to an undesirable increase of training time (the red triangle). Our dynamic strategy achieves a tradeoff by increasing a portion of GPU memory and maintains a nearly identical time cost with original Caffe. Particularly, among those seven neural networks with incremental depth, the swapping strategy is more beneficial to UDNN, as ResNet-152, almost 64\% involved feature maps are offloaded to host memory and only 8.3GB GPU memory is required, making training UDNNs on most of the mid to low-end GPUs possible.

\textbf{\emph{B}. Efficiency-optimal minibatch size}. Given 12GB GPU memory budget, the efficiency-optimal minibatch size after applying the memory optimization technique increases by 22-315\% and keeps almost the same throughput with original Caffe, as shown in Table 1. Likewise, the improvement for UDNN is more notable than other mid-depth networks on account of dramatical memory saving from network-wide to layer-wide. Distinctively, the training throughput grows 17\% and 26\% for ResNet-101 and ResNet-152 respectively because the Caffe's memory-restricted minibatch size is not enough to invoke all the GPU computing resources.

\textbf{\emph{C}. Scaling to distributed cluster}. We implement a parameter server architecture on Caffe to scale training process across the distributed GPU cluster. The parameter server are implemented in the host-side on the same server where GPU worker is equipped and each server takes charge of one parameter shard and stores in the host memory. P2P communication among cluster is implemented using sockets. The most common data parallelism and the most effective BSP are adopted. With a constant number of epochs, larger minibatch size is capable of reducing the times of parameter sychronization, which means, lower communication overhead across cluster makes the whole system center much more on computation.

\begin{table}
\caption{Effectiveness of minibatch size selection}
\resizebox{86mm}{!}{
\begin{tabular}{|l|>{\centering}p{2cm}|>{\centering}p{1.8cm}|>{\centering}p{1.6cm}|>{\centering}p{1.8cm}|>{\centering}p{1.7cm}|}
\hline
\multirow{2}{*}{} & \multicolumn{2}{c|}{AccUDNN} & \multicolumn{2}{c|}{Caffe} & \multirow{2}{1.7cm}{Minibatch size (+\%)}\tabularnewline
\cline{2-5} \cline{3-5} \cline{4-5} \cline{5-5}
 & Efficiency-optimal minibatch  & Throughput (images/s) & Max trainable minibatch & Throughput (images/s) & \tabularnewline
\hline
AlexNet & 1032 & 853 & 848 & 865 & \textbf{\color{blue} 22}\tabularnewline
\hline
GoogleNet & 286 & 245 & 150 & 263 & \textbf{\color{blue} 91}\tabularnewline
\hline
VGG-16 & 192 & 50 & 80 & 51 & \textbf{\color{blue} 140}\tabularnewline
\hline
ResNet-34 & 276 & 99 & 91 & 99 & \textbf{\color{blue} 203}\tabularnewline
\hline
ResNet-50 & 112 & 66 & 33 & 66 & \textbf{\color{blue} 239}\tabularnewline
\hline
ResNet-101 & 74 & \emph{\textbf{34}} & 20 & \emph{\textbf{29}} & \textbf{\color{blue} 270}\tabularnewline
\hline
ResNet-152 & 54 & \emph{\textbf{24}} & 13 & \emph{\textbf{19}} & \textbf{\color{blue} 315}\tabularnewline
\hline
\end{tabular}
}
\end{table}

\begin{figure}[ht]
\centering
\includegraphics[scale=0.4]{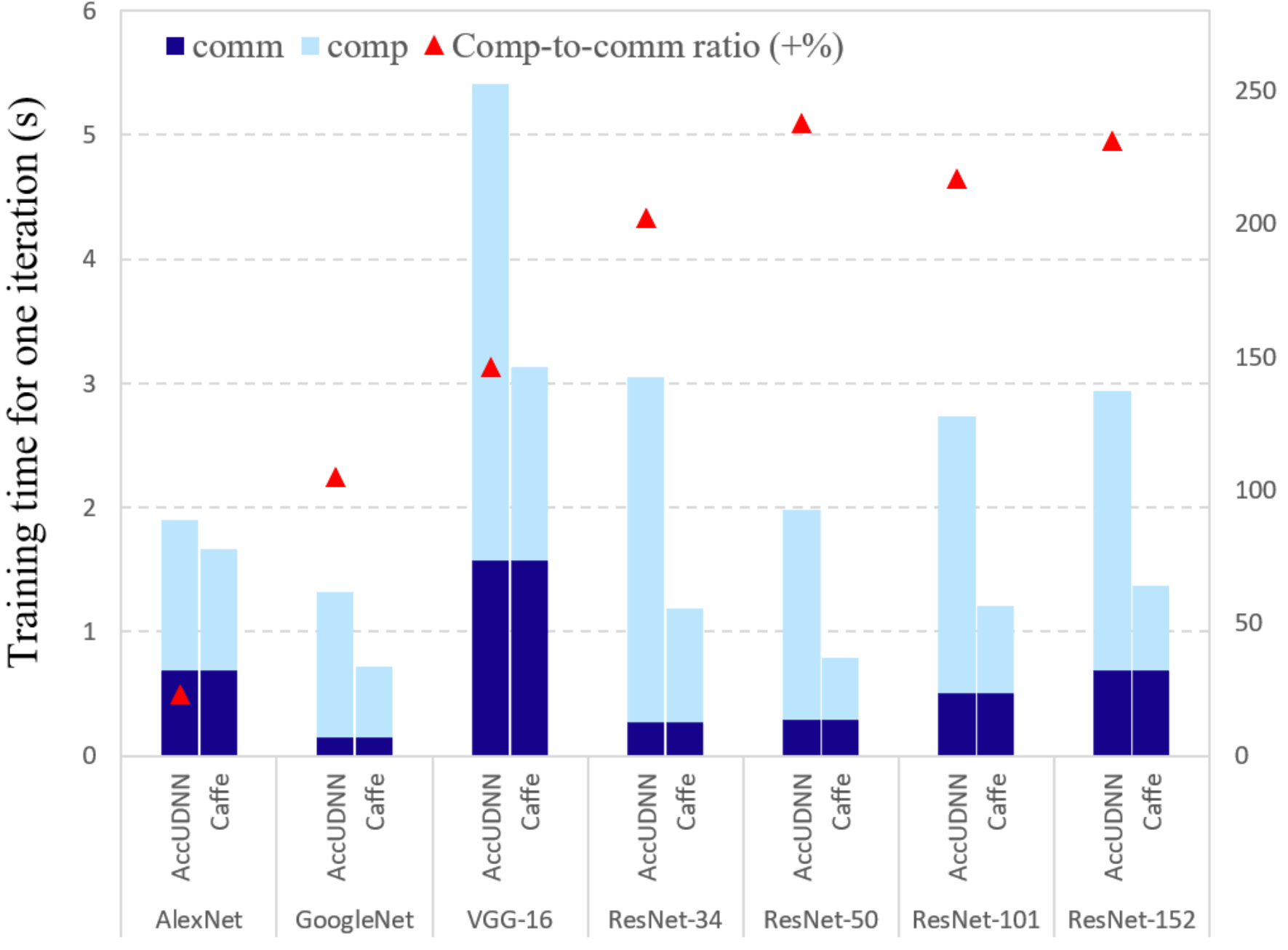}
\caption{Performance in distributed cluster}
\label{fig:comp_to_comm}
\end{figure}

In Figure~\ref{fig:comp_to_comm}, with 8 servers (one server one GPU), by deploying AccUDNN accelerator on individual GPU, we set minibatch size as the efficiency-optimal one in Table 1, finally, the computation-to-communication ratio(the red triangle) of whole system stands in stark contrast to Caffe-PS. Accordingly, the training time is saved about 48 percentage for ResNet-152. Moreover, the AccUDNN is also appropriate for those neural networks with a mass of model parameters, as well as heavy communication overhead, like VGG.

\textbf{\emph{D}. Acceleration for UDNN}. For ResNet-101, our scheme achieves 7.2x speed-up with 8 GPUs, surpassing Caffe-PS apparently. Due to underutilization of GPU computing resource in Caffe default mode, AccUDNN gets the benefit from both the efficiency improvement on single GPU and reduction of communication overhead among cluster. Without any communication optimization, we have achieved a satisfied scaling efficiency. Details list in Table 2. According to our learning rate rules, we set learning rate as 0.22 for total minibatch size 592 and meet final accuracy 76.2\%. AccUDNN performs better on ResNet-152.

\begin{table}
\caption{Acceleration for UDNN ResNet-101/152}
\resizebox{86mm}{!}{
\begin{tabular}{|l|l|c|>{\centering}p{2cm}|>{\centering}p{1.8cm}|>{\centering}p{1.6cm}|}
\hline
 &  & GPUs & Minibatch size & Throughput (images/s) & Scaling efficiency\tabularnewline
\hline
\multirow{3}{*}{Resnet-101} & Caffe & 1 & 20 & 30 & 1\tabularnewline
\cline{2-6} \cline{3-6} \cline{4-6} \cline{5-6} \cline{6-6}
 & Caffe\_PS & 8 & 20 {*} 8 = 160 & 132 & \textbf{\color{blue} 4.4}\tabularnewline
\cline{2-6} \cline{3-6} \cline{4-6} \cline{5-6} \cline{6-6}
 & AccUDNN\_PS & 8 & 74 {*} 8 = 592 & 217 & \textbf{\color{blue} 7.2}\tabularnewline
\hline
\multirow{3}{*}{Resnet-152} & Caffe & 1 & 13 & 19 & 1\tabularnewline
\cline{2-6} \cline{3-6} \cline{4-6} \cline{5-6} \cline{6-6}
 & Caffe\_PS & 8 & 13 {*} 8 = 104 & 76 & \textbf{\color{blue} 4}\tabularnewline
\cline{2-6} \cline{3-6} \cline{4-6} \cline{5-6} \cline{6-6}
 & AccUDNN\_PS & 8 & 54 {*} 8 = 432 & 147 & \textbf{\color{blue} 7.7}\tabularnewline
\hline
\end{tabular}
}
\end{table}

\begin{figure}[ht]
\centering
\includegraphics[scale=0.36]{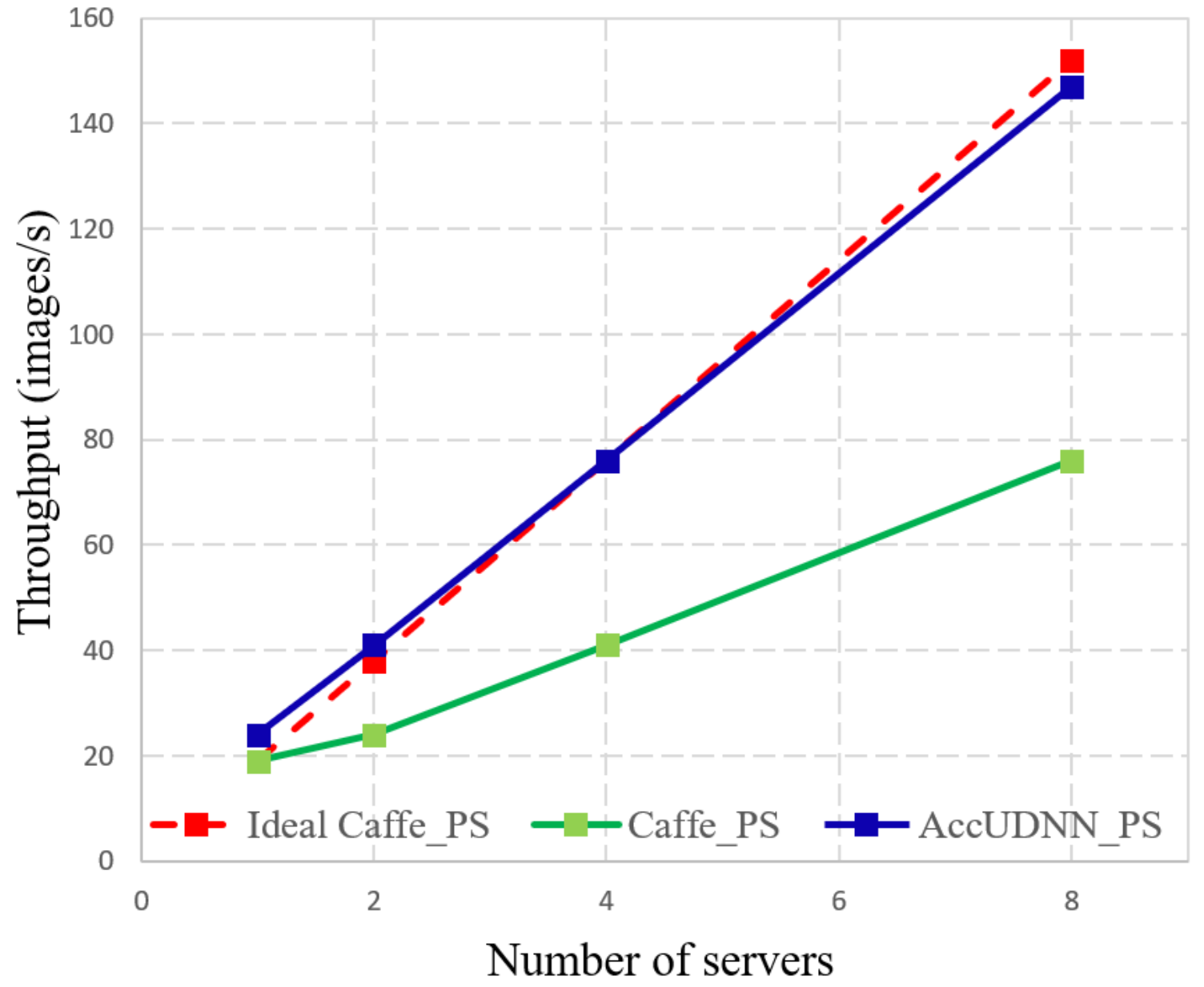}
\caption{Throughput scalability of ResNet-152}
\label{fig:scaling_efficiency}
\end{figure}

Figure~\ref{fig:scaling_efficiency} also exhibits the throughput scalability of ResNet-152 under different cluster sizes. Caffe-PS gives a diminishing return when more GPUs putting into use, lagging far behind the ideal scalability. It is because the computation-to-communication ratio gets lower, accordingly, a larger portion of the training time is spent to deal with the parameter synchronization among cluster. AccUDNN\_PS surpasses Caffe\_PS distinctly owing to better utilization of GPU computing resources and fewer parameter synchronization. Besides, AccUDNN\_PS scales near linearly with the ideal condition of Caffe\_PS, which can be explained as the much more bigger computation-to-communication ratio is capable to cover the small-scale increment of communication overhead for small and medium-sized cluster. Furthermore, the experiment also validates that adopting larger minibatch size for every single GPU, reducing times of parameter synchronization, is the most straightforward and effective way to relieve the cost of communication among distributed cluster in comparison with those lossy communication optimization techniques.

\textbf{\emph{E}. Comparison with Related works}. Recent work SuperNeurons\cite{wang2018superneurons} couples the swapping technique with recomputation mechanism to tackle the trainability issues of UDNN. They finally achieve a GPU memory-optimal state and enable much larger minibatch size to be processed, but the performance, i.e., training efficiency usually discounts when realizing memory-optimal. This can be observed from Table 5 and Figure 14 in SuperNeurons's original paper. The improvement of AccUDNN lies in the shift from memory-optimal to efficiency-optimal, we dedicate to make the memory optimization technique truly benefits the training process, not limited to memory itself.


For a fair comparison, we also test SuperNeurons on our platform and obtain the results in Table 3. The efficiency-optimal minibatch size achieved by AccUDNN is 54-300\% larger than SuperNeurons's and the corresponding throughput also increase 10-106\%. We believe the performance difference of SuperNeurons is caused by the inevitable overhead of recomputation, especially when the minibatch size is relatively big for AlexNet.

Another related work is vDNN\cite{rhu2016vdnn}. We also observe that vDNN lags AccUDNN with 11\% performance loss when training VGG16 with minibatch size 192. In addition, AccUDNN is an integrated solution that can automatically explore the efficiency-optimal hyperparameter, not limited to memory optimization, which is not involved in both vDNN and SuperNeurons.

\textbf{\emph{F}. The overhead of AccUDNN}. We evaluate the performance overhead of AccUDNN in this part. Because of the elimination of overhead in the training process, AccUDNN's overhead mainly focuses on the early modelling and joint solution of efficiency-optimal hyperparameter and its matched swapping strategy.

First is the information gathering stage in memory optimizer. The operations of static information analyzer include simple graph traverse and basic arithmetical operation without running the actual network. Dynamic information profiler only samples a single iteration in the early five epoches respectively while these iterations are still contributory to the training process. Second is the solving process in hyperparameter tuner. Algorithm 2 has \emph{O}(\emph{N}$^{2}$) time complexity. In general, the one-off overhead of AccUDNN is on the order of a few hundreds of seconds, which is negligible relative to the days of the training process, especially the time saved for UDNN.

\begin{table}
\caption{Comparison with SuperNeurons}
\resizebox{86mm}{!}{
\begin{tabular}{|l|c|c|c|c|c|c|}
\hline
\multirow{2}{*}{} & \multicolumn{3}{c|}{Efficiency-optimal minibatch size} & \multicolumn{3}{c|}{Corresponding throughput}\tabularnewline
\cline{2-7} \cline{3-7} \cline{4-7} \cline{5-7} \cline{6-7} \cline{7-7}
 & SuperNeurons & AccUDNN & (+\%) & SuperNeurons & AccUDNN & (+\%)\tabularnewline
\hline
AlexNet & 640 & 1032 & \textbf{\color{blue}61} & 414 & 853 & \textbf{\color{blue}106}\tabularnewline
\hline
VGG-16 & 48 & 192 & \textbf{\color{blue}300} & 30 & 50 & \textbf{\color{blue}67}\tabularnewline
\hline
ResNet-50 & 48 & 112 & \textbf{\color{blue}133} & 60 & 66 & \textbf{\color{blue}10}\tabularnewline
\hline
ResNet-101 & 48 & 74 & \textbf{\color{blue}54} & 30 & 34 & \textbf{\color{blue}13}\tabularnewline
\hline
ResNet-152 & 32 & 54 & \textbf{\color{blue}68} & 17 & 24 & \textbf{\color{blue}42}\tabularnewline
\hline
\end{tabular}
}
\end{table}

\section{Related work}
Existing systematic approaches to overcome the GPU memory shortage for UDNN are: 1) model parallelism, i.e., split the UDNN across multiple GPUs, 2) optimize memory usage on single GPU.

Model parallelism provides a straightforward solution for training UDNN by splitting the whole UDNN across multiple GPUs(maybe among same or different servers). From Distbelief\cite{Dean2013Large} to Tensorflow\cite{abadi2016tensorflow}, they all support the model parallelism mode. However, on account of the data dependencies, model parallelism demands intra-network even intra-layer synchronization frenquently. The excessive communication, especially among different servers, deteriorates training efficiency severely, as pointed in \cite{Coates2013Deep}, when training a network with 1.3 billion parameters from 36 GPUs to 64 GPUs, the efficiency penalty reaches up to 40\%. Thus, model parallelism is not widely used in practice.

Memory optimization techniques for single GPU include recomputation\cite{chen2016training,gruslys2016memory}, data encode compression\cite{Jain2018Gist}, swap out/in strategy between GPU and host memory\cite{rhu2016vdnn,meng2017training, le2018tflms,cui2016geeps}. Chen\cite{chen2016training} proposes the idea of trading computation for memory, dropping partial feature maps in forward propagation and recomputes them when needed in backward propagation. Gist\cite{Jain2018Gist} takes a slightly different approach, they store those feature maps in an encoded representation and decode them when used again. vDNN\cite{rhu2016vdnn} utilizes the relatively abundant host memory as backing store for GPU and smartly swap feature maps out to host side. In addition, \cite{wang2018superneurons} makes a combination of recomputation and swap operation, achieving a greater degree of memory reduction. These techniques do reduce GPU memory consumption, but they all give rise to extra performance overhead from 4-30\%.

For hyperparameter minibatch size selection, increasing interest has turned to large minibatch size to speed up the training process\cite{smith2017don, akiba2017extremely, seide2014parallelizability}. For a single GPU, large minibatch size enables higher parallelism of GPU computing units\cite{chetlur2014cudnn}. In the distributed cluster, large minibatch size is conductive to scale data parallelism mode across multiple GPUs. Two avenues to increase minibatch size are: use more GPUs and increase minibatch size on per GPU respectively.

The former avenue has already explored by IT companies with abundant hardware resources. \cite{goyal2017accurate} Facebook's experiment employs 256 GPUs to train ResNet-50 with a minibatch size of 8192 and finishes in 1 hour. \cite{akiba2017extremely} extends Facebook's experiment to 1024 GPUs and finishes in 15 minutes. However, the later avenue hasn't been fully studied due to the limitation of GPU memory capacity, in the above-mentioned two experiments, minibatch size on per GPU is set as 32 only. Given a fixed number of servers, the smaller minibatch size processed by each server is, the more frequent global parameter synchronization among cluster is. With common commodity Ethernet 100x slower than physical memory, the time cost of inter-machine communication among dozens or hundreds of servers usually leads to diminishing returns.

\section{Conclusion}
In this paper, we propose AccUDNN to accelerate the training process of UDNN from the perspective of GPU memory optimization. By applying the performance-model guided dynamic swap out/in strategy, memory optimizer mitigates those suitable feature maps to host memory, thereby breaking down the trainability restriction in a way without incurring performance degradation. Then, a hyperparameter tuner is adopted to explore the efficiency-optimal hyperparameter setting after applying the memory optimization technique so as to further improve the training efficiency. Finally, evaluations against state-of-the-art DL framework Caffe have demonstrated the effectiveness and efficiency of AccUDNN.

\bibliographystyle{plain}
\bibliography{\jobname}

\end{document}